\documentclass[runningheads]{llncs}
\usepackage[T1]{fontenc}
\usepackage[english]{babel}
\usepackage{amsmath}
\usepackage{graphicx}
\usepackage{subcaption} 
\usepackage[colorlinks=true, allcolors=blue]{hyperref}
\usepackage[style=numeric]{biblatex}
\usepackage{color}

\begin{document}
\title{Neural Cellular Automata Learn General Features in their Hidden Channels}
\author{Etienne Guichard\inst{1}\orcidID{0009-0005-5300-8182} \and
Stefano Nichele\inst{1}\orcidID{0000-0003-4696-9872}}
\authorrunning{E. Guichard \and S. Nichele}
\institute{Østfold University of Applied Sciences, BRA Veien 4, 1757 Halden, Norway \\
\email{\{etienne.guichard, stefano.nichele\}@hiof.no}}
\maketitle              
\begin{abstract}
Modern deep learning models achieve impressive generalization through over-parameterization, but this paradigm often struggles with overfitting and memorization in few-shot regimes. Neural Cellular Automata (NCAs) offer a highly parameter-efficient alternative, yet research has focused primarily on their output, leaving the role of their internal hidden channels largely unexplored. In this paper, we investigate the internal dynamics of NCA hidden channels and introduce a novel transfer-learning mechanism that injects a pretrained teacher's hidden states into a student model to guide early optimization. Evaluated on few-shot and scale-variant MNIST benchmarks, NCAs outperform comparable recurrent and feed-forward architectures, demonstrating superior generalization with a minimal parameter budget ($\approx$ 9,800 parameters). Mechanistic analysis reveals that the hidden channels decouple feature extraction from uniform classification consensus by absorbing morphological complexity and converging to mutually orthogonal states. Furthermore, we demonstrate that these hidden channels capture general, scale-invariant topological primitives rather than class-specific templates. This allows a student model to achieve strong few-shot performance on unseen classes using features transferred from a teacher trained only on a subset of digits (0--5). Our results highlight the potential of utilizing hidden-state dynamics as a robust, decentralized computational substrate for parameter-efficient transfer learning.

\end{abstract}
\section{Introduction}

Modern deep learning relies heavily on over-parameterization to achieve state-of-the-art performance across a myriad of domains \cite{nakkiran2020deep, belkin2019reconciling}. Empirical scaling laws suggest a power-law relationship in which continuous increases in parameter counts necessitate commensurate increases in both training data and computational resources \cite{kaplan2020scaling, hoffmann2022training}. While this paradigm has yielded exceptional performance in complex tasks where massive, high-quality datasets are readily available \cite{brown2020language}, it presents a fundamental challenge for few-shot learning \cite{wang2020generalizing}. In strictly data-limited environments, massively over-parameterized models are highly susceptible to overfitting, often tending to memorize the sparse training dataset rather than learning generalizable abstractions \cite{carlini2021extracting, feldman2020does}. Consequently, the community has increasingly sought parameter-efficient architectures and strong inductive biases capable of robust generalization without requiring exhaustive data or computational budgets \cite{finn2017model, snell2017prototypical}.

Neural Cellular Automata (NCAs) represent a promising alternative, a class of decentralized machine learning models that may offer a compelling solution to the generalization bottlenecks of few-shot learning. Inspired by the biological processes of cell morphogenesis and self-organization \cite{turing1952chemical, levin2012morphogenetic, nichele2017deep}, NCAs operate by repeatedly applying highly localized, shared update rules across a spatial grid. Initially popularized by Mordvintsev et al. \cite{mordvintsev2020growing} for image generation and regeneration tasks, NCAs have recently demonstrated remarkable computational capabilities across various domains \cite{sudhakaran2021growing, earle2022illuminating, xu2025neuralcellularautomataarcagi, guichard2025arcncadevelopmentalsolutionsabstraction}. Crucially, despite possessing drastically lower parameter counts—often magnitudes smaller than traditional convolutional or feed-forward networks—NCAs exhibit emergent, complex behaviors and strong robustness \cite{variengien2021towards, gilpin2019cellular}. Their strict adherence to local message-passing acts as a structural inductive bias, effectively preventing the memorization of global spatial coordinates and forcing the network to learn robust, coordinate-free structural rules \cite{randazzo2020self}.

Despite these advantages, the majority of existing NCA research has predominantly focused on their external capabilities, evaluating models almost exclusively by the values of their output channels \cite{palm2022variational, niklasson2021textures}. Studies frequently assess an NCA's ability to generate specific textures, segment images, or classify inputs based solely on the final convergence of these visible states \cite{walker2022physical}. In contrast, relatively few papers have rigorously investigated the internal dynamics of the NCA's hidden channels \cite{Cisneros_2020}. These hidden channels—analogous to the internal chemical gradients of biological cells—act as a decentralized, dynamic computational substrate. Understanding how information is processed, spatially distributed, and geometrically represented within this hidden space remains an open challenge critical for interpreting the model's emergent capabilities.

Traditional transfer learning relies on the extraction and fine-tuning of pre-trained synaptic weights \cite{yosinski2014how, zhuang2020comprehensivesurveytransferlearning}. In this paper, we explore a different method exclusive to NCAs: transfer learning via hidden channels rather than weights. Because an NCA's hidden channels store distributed, spatial representations of features across developmental steps, and are incorporated into its computational graph, we hypothesize that injecting a ``teacher'' NCA's mature hidden state into a ``student'' NCA can serve as a form of \textit{bootstrapping}. By bypassing the chaotic initial phases of state formation, a student network can immediately leverage well-structured topological primitives—such as edges and intersections—without requiring identical synaptic parameters or architecture, paving the way for highly efficient transfer.

\section{Experimental Setup}
\subsection{Generalisation Experiments}\label{section:gen}
The experimental setup for this experiment is as follows: we test an NCA against other models (mainly a Globally perceptive recurrent model, and two variants of a feed-forward model), all models have the same parameter count and are trained on $k$ examples per class (on the MNIST training dataset) to see how well they generalize with few examples. Each model is trained for 3,000 iterations using the $AdamW$ optimizer with a learning rate of $1e-3$ and is finally evaluated on the entire MNIST test set. The NCA model is evaluated with MSE loss, while the other models (producing logits) are evaluated using cross-entropy loss. To further test generalization, we also evaluate the model on scaled-down versions of the MNIST dataset, where MNIST digits are linearly scaled down (using a nearest-neighbor downscaler) and padded back up to $28 \times 28$ images with background pixels. All experiments were conducted on 100 random seeds. We also set an 80\% threshold (a reasonably good score) as the metric for the model's ability to generalize, since with more data points, the models converge faster, with the idea being that at this lower score, we should be able to distinguish between models better.

\subsection{Inductive Bias}\label{sec:id}
\begin{figure}[htbp]
\centering
\includegraphics[width=1.0\linewidth]{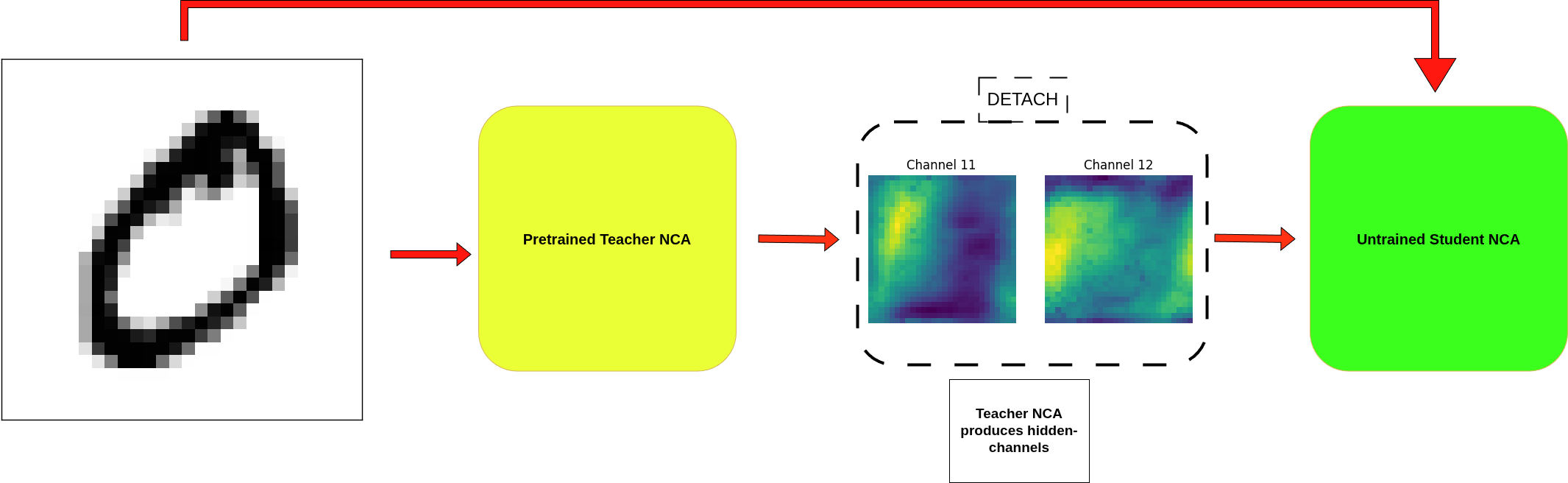}
\caption{\label{fig:dial1} Diagram showing how the hidden channels are extracted from a teacher NCA and used in a Student NCA. The teacher NCA sees a digit, produces a classification, and some hidden channels are used in computing. These hidden channels are extracted and detached and passed to the student NCA alongside the digit before the forward pass in training.}
\end{figure}

We repeat the same experiment as in \ref{section:gen}, with one caveat: during training, for the first 1500 steps of backpropagation, we inject the non-solution hidden state of an NCA pretrained on the entire MNIST dataset and has seen the $k$ examples we are training on. By 'non-solution,' we mean distinguishing the NCA's different channels. Channel 0 is used to inject the MNIST digit, channels 1-10 are used for one-hot classification, and channels 11-15 are used for computation; these are the non-solution channels. From steps 1500-3000 of backpropagation, we remove the hidden-state injection; the NCA has to learn to reconstruct this hidden state on its own. We ensure that no gradient information from the "teacher" NCA is available to the "student" NCA by detaching the injected tensor from the computational graph before training the student. The injected hidden state is only injected at the student's NCA developmental step $T = 0$, see Figure \ref{fig:dial1} for more details. Everything else is as before. During testing, the NCA has to reconstruct the hidden channels itself, without aid from the injection model.

\subsection{Transfer learning}\label{sec:tf}
Here, we repeat the same experiment as in \ref{sec:id}, but the injected state from the pretrained model is always present throughout training and testing, injected only at the student's NCA developmental step $T = 0$. This effectively turns the NCA into a conditioned readout model, where it learns to identify the digit and the hidden state to produce the one-hot encoding.

\section{Experimental Results}\label{sec:res}
\subsection{NCA Generalisation}\label{sec:generalization}
\begin{figure}[htbp]
\centering
\includegraphics[width=1.0\linewidth]{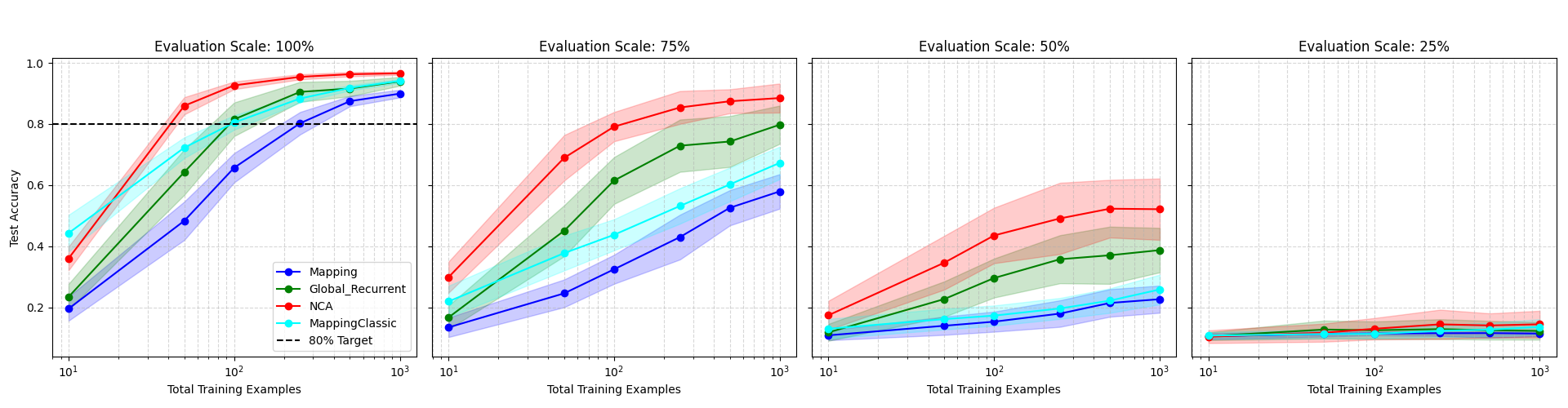}
\caption{\label{fig:tst1}All 4 models, tested on $k \in \{1,5,10,25,50,100\}$ example per class, all have $\approx 9800$ parameters. Since there are $10$ classes, $10^1$ examples equals one example per class. Scales range from 100 percent MNIST to 25 percent MNIST, with 25 percent intervals.}
\end{figure}

As shown in Figure \ref{fig:tst1}, the NCA outperforms all other models in K-shot generalization, except for the mapping classic in 1-shot experiments. The model also crosses the 80\% threshold considerably earlier than all other models at about 4 examples per class, as opposed to 10 (for mapping classic and Global Recurrent) and 11 (for Mapping). When evaluated on different scales, the NCA outperforms all models across all k-shot tests at 75\% and 50\%; at 25\%, it is not statistically distinguishable from other models. 

\subsection{Inductive Bias}\label{sec:ib}
\begin{figure}[htbp]
\centering
\includegraphics[width=1.0\linewidth]{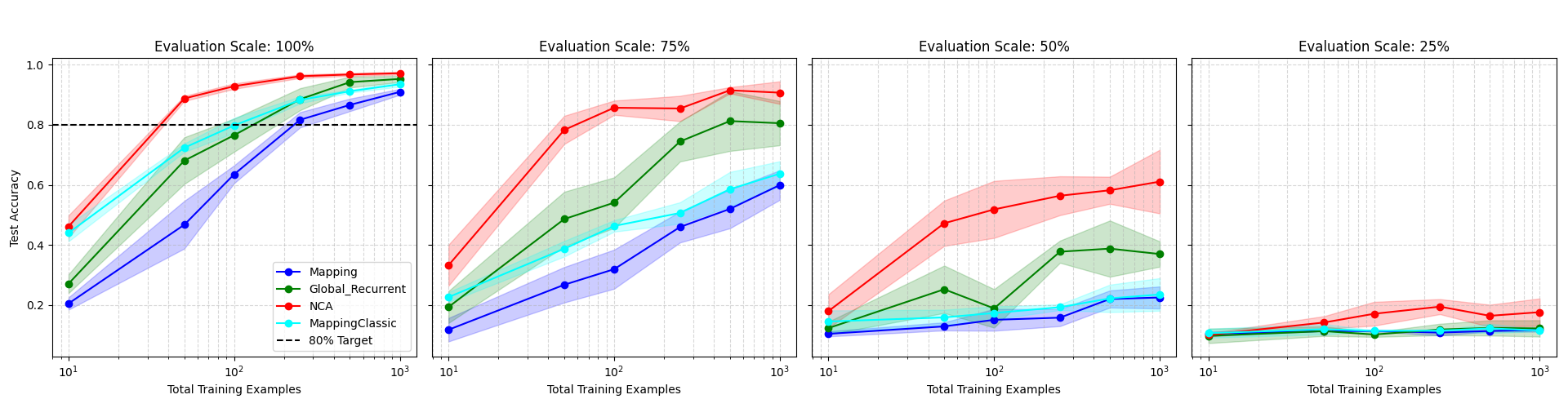}
\caption{\label{fig:tst3}All 4 models, tested on $k \in \{1,5,10,25,50,100\}$ example per class, all have $\approx 9800$ parameters. Since there are $10$ classes, $10^1$ examples equals one example per class. Scales range from 100 percent MNIST to 25 percent MNIST, with 25 percent intervals. NCA receives a non-solution hidden state across a portion of training.}
\end{figure}

As shown in Figure \ref{fig:tst3}, when allowed to see useful hidden channel representations across a portion of the K-shot learning regime, the NCA outperforms all models on all K-shot learning problems, crossing the 80\% threshold somewhere between 3 and 4 examples. The NCA also outperforms all models in scaled-down evaluations, and surprisingly achieves a statistically significant (yet modest) result on even the 25\% scale, something it was not capable of without this mechanism.  

\begin{figure}[htbp]
\centering
\includegraphics[width=0.7\linewidth]{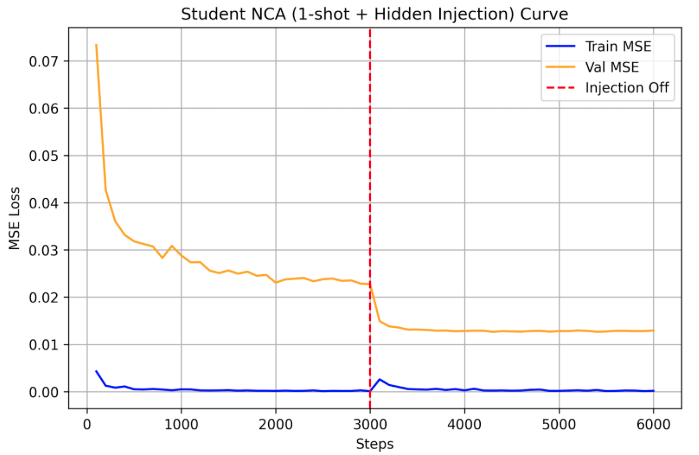}
\caption{\label{fig:tst3.1} Training loss (blue) and validation loss (orange) of one run of the NCA in a 1-shot environment trained with the hidden injections.}
\end{figure}

Figure \ref{fig:tst3.1} shows the training loss (blue) and validation loss (orange) of a single run of a student NCA in a one-shot environment. The dotted red line represents when the injections of the hidden state are no longer provided. As shown in the figure, training loss quickly reaches a plateau (of near 0) while validation loss also stabilizes as the injection is provided. Once the injection is no longer provided, training loss briefly spikes, and validation loss drops considerably. This has mostly to do with the difference in training and testing before the injections are removed. During training, the NCA is already given well-constructed hidden channels, while during testing, it has to construct them itself, a task it has not been trained on yet. 

\subsection{Transfer Learning}\label{sec:inductive_bias}
\begin{figure}[htbp]
\centering
\includegraphics[width=1.0\linewidth]{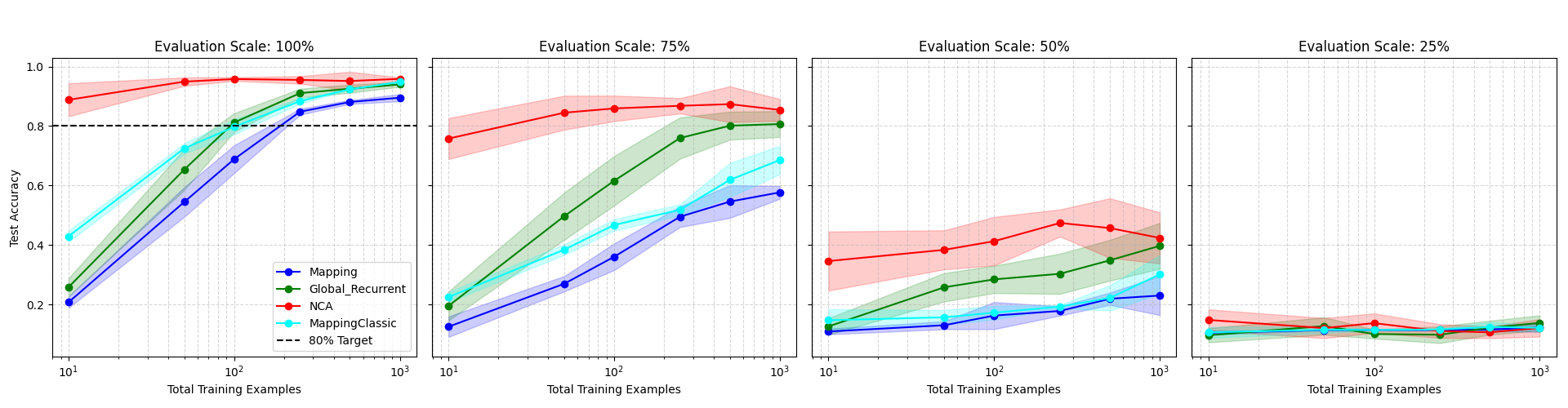}
\caption{\label{fig:tst4} All 4 models, tested on $k \in \{1,5,10,25,50,100\}$ example per class, all have $\approx 9800$ parameters. Since there are $10$ classes, $10^1$ examples equals one example per class. Scales range from 100 percent MNIST to 25 percent MNIST, with 25 percent intervals. NCA receives the non-solution hidden channels across all of training and validation.}
\end{figure}

As shown in Figure \ref{fig:tst4}, when the student NCA is allowed to see the hidden channels throughout both training and testing, the performance on K-shot learning vastly outperforms all other models. 

\subsection{Reduced Knowledge Transfer Learning}
Here we perform the same experiment as before, but the pretrained NCA only sees digits 0--5, while the Decoder NCA is trained on K-shot generalization on all MNIST digits.

\begin{figure}[htbp]
\centering
\includegraphics[width=1.0\linewidth]{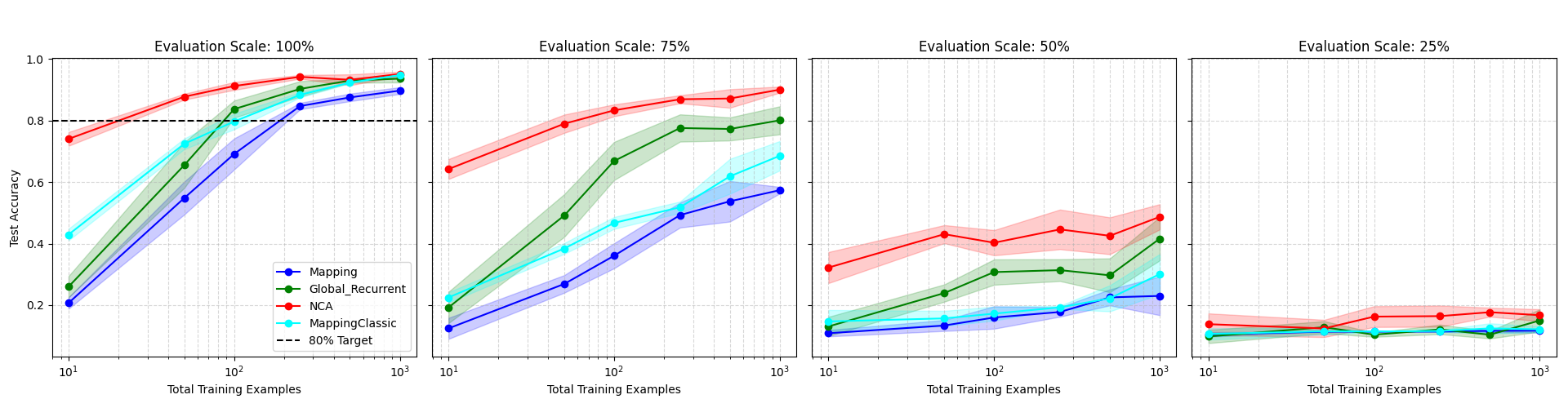}
\caption{\label{fig:tst5} All 4 models, tested on $k \in \{1,5,10,25,50,100\}$ example per class, all have $\approx 9800$ parameters. Since there are $10$ classes, $10^1$ examples equals one example per class. Scales range from 100 percent MNIST to 25 percent MNIST, with 25 percent intervals. NCA receives the non-solution hidden channels across all of training and validation.}
\end{figure}

As shown in Figure \ref{fig:tst5}, even when the teacher NCA is trained on a 5-way MNIST dataset, the features extracted are sufficient for the student model to achieve good generalization in 1-100-shot learning. The drop-off in evaluation-scale performance is also similar in magnitude to that of a teacher NCA trained on the entire MNIST dataset. 

\section{Hidden Channel Analysis}
While the experimental results in Section \ref{sec:res} demonstrate the NCA's capability in few-shot generalization and transfer learning, they do not explain the underlying mechanisms driving this performance. Unlike standard feed-forward networks, where intermediate activations represent static feature hierarchies, the hidden channels of an NCA (channels 11–15) act as a decentralized, dynamic computational substrate, capable of storing representations as well. In this section, we aim to probe the internal dynamics of these hidden channels to understand how NCAs leverage the development of their hidden states for classification. The models represented in this analysis are a fully trained teacher NCA (akin to an NCA from Section \ref{sec:generalization} trained on all of MNIST) and a student NCA (such as that in Section \ref{sec:inductive_bias}) trained using injected inputs from the teacher NCA on a 1-shot MNIST problem. 

\subsection{Scale Free Representation}\label{sec:scale_free}
\begin{figure}[htbp]
\centering
\includegraphics[width=1.0\linewidth]{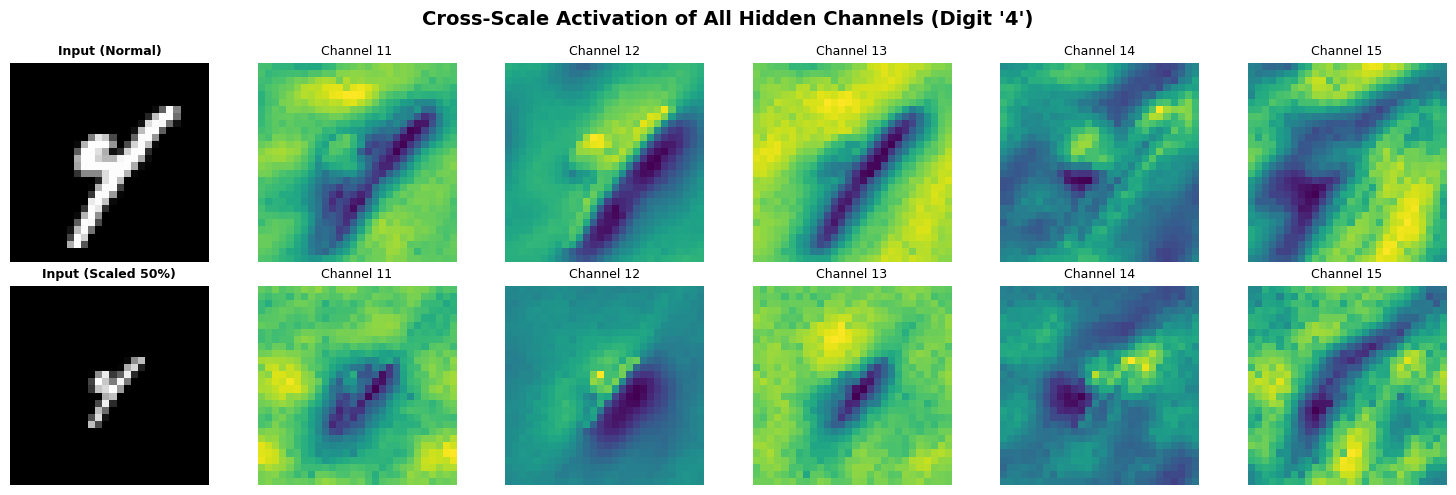}
\caption{\label{fig:4sc} \textbf{Top:} Hidden channel representations built by the NCA on a full-size MNIST 4. \textbf{Bottom:} Hidden channel representations built by the same NCA on a 50\% scale version of the same 4.}
\end{figure}

As shown in Figures \ref{fig:tst1}, \ref{fig:tst3}, \ref{fig:tst4}, and \ref{fig:tst5}, the NCA model generally outperformed all other models on varying scales of the MNIST data set. Figure \ref{fig:4sc} hints at the underlying mechanism. As can be seen, the broad features the NCA builds in its hidden channel, such as topological intersections (channel 14) and edges (channel 12) are scale invariant. Because the NCA operates exclusively via local neighborhood communication, it is incapable of memorizing global spatial coordinates and features. Consequently, the hidden channels are forced to learn scale-invariant structures rather than pixel-level patterns.

\subsection{Temporal Evolution of Spatial Variance}\label{sec:spatial_variance}
\begin{figure}[htbp]
    \centering
    \begin{subfigure}[b]{0.48\textwidth}
        \centering
        \includegraphics[width=\textwidth]{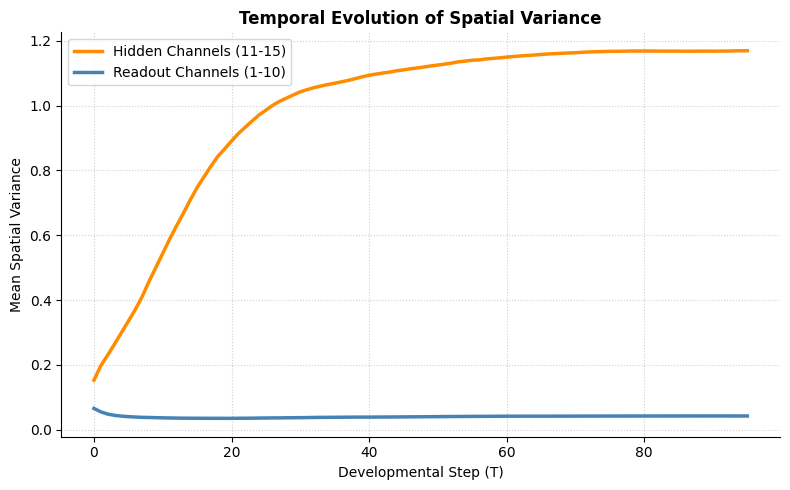}
        \caption{Teacher NCA}
        \label{fig:temepevoteach}
    \end{subfigure}
    \hfill
    \begin{subfigure}[b]{0.48\textwidth}
        \centering
        \includegraphics[width=\textwidth]{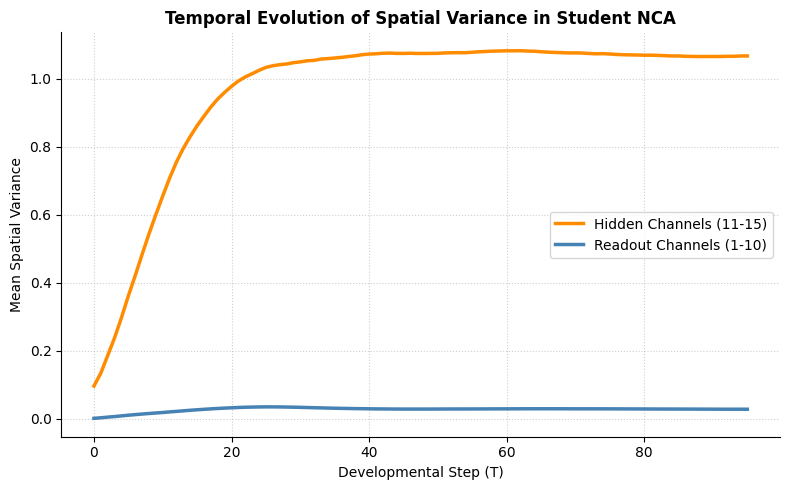}
        \caption{Student NCA}
        \label{fig:tempevostud}
    \end{subfigure}
    \caption{Temporal evolution of the spatial variance of the hidden channels of (a) the teacher NCA and (b) the student NCA.}\label{fig:tempvar}
\end{figure}

To quantify how the NCA separates feature extraction from classification, we tracked the mean spatial variance of the hidden channels versus the readout channels over time (Figure \ref{fig:tempvar}).

We compute the mean spatial variance $V_t(\mathcal{C})$ for a specific subset of channels $\mathcal{C}$ at developmental step $t$. Let $x^{(i)}_{t, c, h, w}$ represent the activation value for the $i$-th batch sample at channel $c$ and spatial coordinates $(h, w)$. The temporal evolution of spatial variance is defined as:

\begin{equation}
    V_t(\mathcal{C}) = \frac{1}{N |\mathcal{C}|} \sum_{i=1}^{N} \sum_{c \in \mathcal{C}} \left[ \frac{1}{H W} \sum_{h=1}^{H} \sum_{w=1}^{W} \left( x^{(i)}_{t, c, h, w} - \mu^{(i)}_{t, c} \right)^2 \right]
    \label{eq:spatial_variance}
\end{equation}

where $N$ is the batch size, $H$ and $W$ are the spatial dimensions of the grid, and $\mu^{(i)}_{t, c}$ is the spatial mean of that specific channel map, given by:

\begin{equation}
    \mu^{(i)}_{t, c} = \frac{1}{H W} \sum_{h=1}^{H} \sum_{w=1}^{W} x^{(i)}_{t, c, h, w}
\end{equation}

In our analysis, we evaluate this metric separately for the hidden channels ($\mathcal{C}_{\text{hidden}} = \{11, \dots, 15\}$) and the readout channels ($\mathcal{C}_{\text{readout}} = \{1, \dots, 10\}$).

As the NCA processes an input, the hidden channels exhibit a rapid increase in spatial variance, corresponding to the complex morphological formation of features, followed by a plateau, corresponding to the stabilization of the features. Conversely, the classification readout channels maintain a near-zero spatial variance, reflecting a uniform grid-wide consensus. This decoupling demonstrates that the hidden channels actively absorb the task's structural complexity, providing a stabilized representation that enables the readout channels to converge on a uniform classification.

\subsection{Cross-Channel Cosine Similarity Over Time}

\begin{figure}[t]
    \centering
    \begin{subfigure}[b]{0.48\textwidth}
        \centering
        \includegraphics[width=\textwidth]{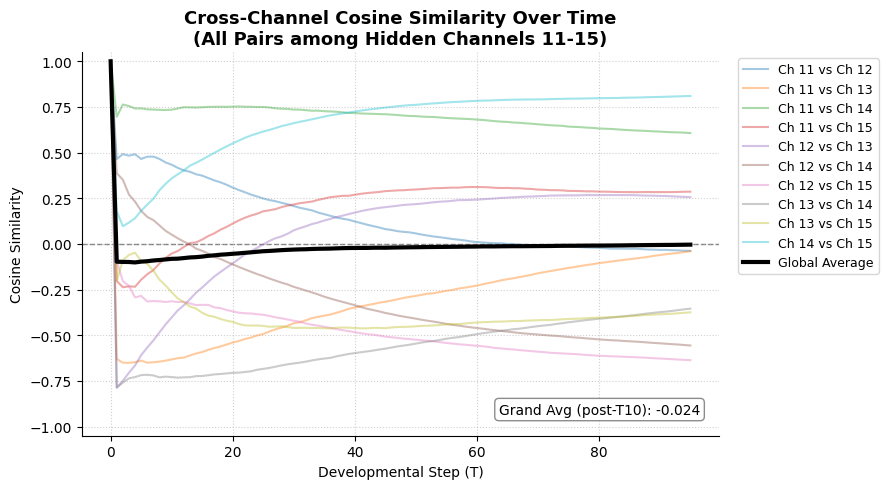}
        \caption{Teacher NCA}
        \label{fig:pca_t1}
    \end{subfigure}
    \hfill
    \begin{subfigure}[b]{0.48\textwidth}
        \centering
        \includegraphics[width=\textwidth]{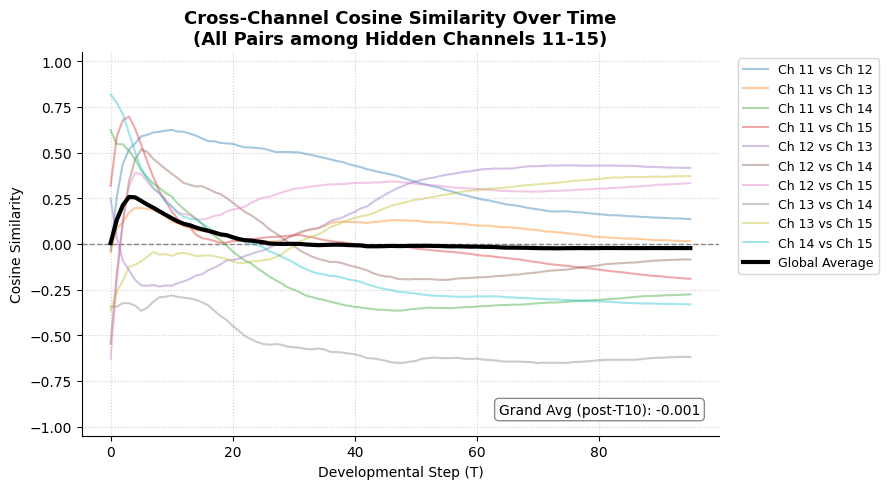}
        \caption{Student NCA}
        \label{fig:pca_t5}
    \end{subfigure}
    \caption{Cross-Channel Cosine Similarity between all channel pairs over time for (a) the teacher NCA and (b) the Student NCA.}\label{fig:cccs}
\end{figure}

Given the restricted parametric capacity of our models ($\approx$ 9,800 parameters), we investigated how the NCA maximizes its representational efficiency. Figure \ref{fig:cccs} illustrates the cross-channel cosine similarity between all pairs of hidden channels over time. Crucially, the global average similarity across all pairs converges to approximately zero. This orthogonality confirms that the NCA eliminates representational redundancy, forcing distinct channels to adopt highly specialized, non-overlapping algorithmic roles (such as synergistic or inhibitory functions). Furthermore, while the pre-trained Teacher model exhibits smooth, pre-orchestrated channel relationships from $T = 0$, the Student model exhibits a brief, high-variance transient phase. This suggests the Student actively translates and reorganizes the injected state to align with its specific parametric topology, rather than passively copying it.

\subsection{Latent Trajectory of Hidden States}\label{sec:latent_trajectory}
\begin{figure}[t]
\centering
\includegraphics[width=0.5\linewidth]{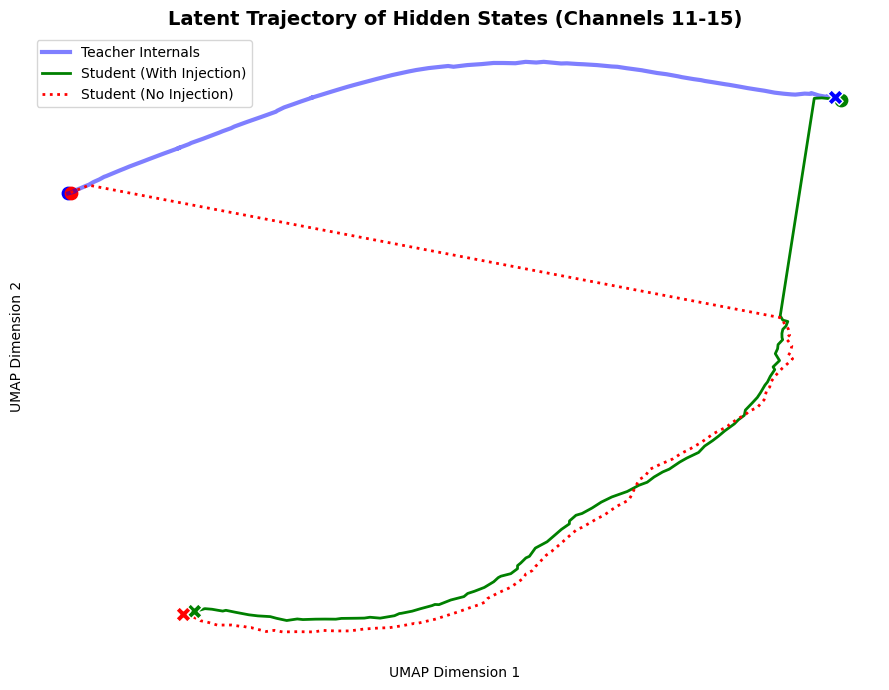}
\caption{\label{fig:latent_traj} 2D UMAP trajectory of the hidden channels for the Teacher (blue), Student with injection (green), and Student with no injection (red).}
\end{figure}

Finally, we visualize the dynamical attractors of the NCA by projecting the continuous temporal trajectory of a single digit's hidden state into a 2D UMAP embedding (Figure \ref{fig:latent_traj}). When evaluating the Student model from a random-state initialization (red trajectory), the system must traverse a chaotic transient manifold before finally settling into its global attractor basin. In contrast, the state-injected Student (green trajectory) initiates its sequence at the exact terminus of the Teacher's mature developmental trajectory. Because the Student possesses distinct weights, it smoothly glides down the gradient of its own loss landscape to its own attractor, entirely bypassing the high-variance phase of morphogenesis. This confirms that hidden-channel injection effectively acts as developmental scaffolding, placing the model directly into a stable, mature representational basin. The fact that the student, even when starting from the teacher's terminal state, follows its own trajectory shows that it is not acting merely as a readout. If this were the case, one would expect the student's hidden channels to have no trajectory at all. 

\section{Discussion}
This study is not intended to prove that NCAs are superior to other machine learning methods; it seeks to shed light on how these models differ and what advantages and disadvantages they entail. Traditionally, the hidden space of NCAs has been seen as a hidden computational storage space, primarily used to pass intermediate messages between local steps until a classification consensus is reached. 

Results from Section \ref{sec:generalization} and Section \ref{sec:inductive_bias} indicate that NCAs may have strong generalization capabilities, particularly in few-shot regimes. Unlike standard feed-forward networks, which often rely on high parameter capacity to fit complex datasets, the recurrent, local nature of NCAs forces them to capture structural patterns with a highly restricted parameter budget. As shown in Section \ref{sec:scale_free}, these representations are scale invariant, helping the NCA generalize further on unseeded data. 

Both the inductive bias and transfer learning experiments show that the hidden channels are not merely useful information storage spaces but also core computational elements that can be exploited. Rather than serving as intermediate information storage channels, they function more as spatially distributed feature maps that the output cells utilize during classification. When we inject a pre-trained teacher's hidden state, we provide the student model with pre-computed feature maps. This bootstrapping allows the student to bypass the chaotic initial phases of state formation and exploit features more readily in few-shot regimes, as visualized in the latent trajectory analysis (Section \ref{sec:latent_trajectory}). We also theorize that the bootstrapping mechanism provides the student with more orthogonal features to learn from, even in few-shot regimes, partially explaining why its performance increases across the entire dataset after seeing only one example.

Our analysis of the internal dynamics, however, reveals a more nuanced division of labor. As shown in the spatial variance analysis (Section \ref{sec:spatial_variance}), the hidden channels capture the complex morphological structure of features, while the readout channels maintain near-zero spatial variance to preserve grid-wide consensus. Furthermore, the near-zero average cosine similarity across channels suggests that the NCA minimizes representational redundancy, forcing distinct channels to adopt highly specialized, orthogonal algorithmic roles.

Perhaps the most notable observation comes from the reduced-knowledge-transfer experiments. The teacher's ability to extract general features from a subset of classes (digits 0--5) that are immediately useful for a student learning unseen classes (digits 6--9) suggests that the hidden channels are capturing fundamental, scale-invariant topological primitives rather than class-specific templates. Because the NCA operates exclusively via local neighborhood communication, it is unable to memorize global spatial coordinates, forcing the network to rely on these robust geometric primitives instead.

Despite these promising qualities, several limitations must be acknowledged:
\begin{itemize}
    \item \textbf{Dataset Complexity:} These experiments were conducted on variants of the MNIST dataset. While MNIST serves as a valuable benchmark for topological and geometric reasoning, it is structurally simple and monochromatic. It remains to be seen whether these scale-invariant features and state-transfer mechanisms scale effectively to high-resolution, complex color datasets (such as CIFAR-10 or ImageNet).
    \item \textbf{Computational Overhead:} Although the NCA achieves high representational efficiency with a low parameter count ($\approx$9,800 parameters), simulating dozens of developmental steps during both training and inference introduces considerable computational and time overhead. Backpropagation through time (BPTT) over many steps remains a scaling bottleneck compared to standard single-pass feed-forward architectures. This also means the comparisons to other models (with the exception of the global recurrent model) do not fully reflect the capability differences, as the recurrent models are afforded a much higher computational budget.
\end{itemize}

\section{Conclusion}
In this study, we investigated the internal representation and transferability of hidden channels in Neural Cellular Automata (NCAs) within data-scarce and scale-variant environments. While traditional research has viewed the hidden space of NCAs merely as intermediate message-passing channels, our work demonstrates a highly coordinated division of labor. Through spatial variance and cross-channel cosine similarity analyses, we showed that the hidden channels actively absorb morphological complexity, allowing the readout channels to converge uniformly on a classification consensus. Crucially, these hidden channels strictly minimize representational redundancy, dynamically self-organizing into mutually orthogonal algorithmic roles over time.

Our experiments on bootstrapping demonstrate that transfer learning can be successfully achieved by injecting pre-computed hidden states rather than solely relying on standard synaptic weight adjustment. This mechanism allows a student model to bypass chaotic initial morphogenesis phases and initialize within a stable, mature attractor basin. Furthermore, our reduced-knowledge transfer experiments confirm that the hidden channels capture scale-invariant topological primitives rather than class-specific templates, enabling robust generalization on unseen classes. 

Despite these promising results, several limitations must be acknowledged. Our evaluation was confined to structurally simple, monochromatic MNIST variants, and the scaling of backpropagation through time (BPTT) over many developmental steps introduces non-trivial computational and time overhead. Future work will focus on scaling these dynamics to higher-resolution color datasets, investigating the behavior of larger multi-channel hidden spaces, and systematically reverse-engineering the orthogonal roles of individual channels to improve the mechanistic interpretability of decentralized learning architectures.

\section*{Acknowledgment}
This work was supported by MishMash - Research Council of Norway, GrantId: 357438.

\printbibliography

@misc{guichard2025arcncadevelopmentalsolutionsabstraction,
      title={ARC-NCA: Towards Developmental Solutions to the Abstraction and Reasoning Corpus}, 
      author={Etienne Guichard and Felix Reimers and Mia Kvalsund and Mikkel Lepperød and Stefano Nichele},
      year={2025},
      eprint={2505.08778},
      archivePrefix={arXiv},
      primaryClass={cs.AI},
      url={https://arxiv.org/abs/2505.08778}, 
}

@misc{xu2025neuralcellularautomataarcagi,
      title={Neural Cellular Automata for ARC-AGI}, 
      author={Kevin Xu and Risto Miikkulainen},
      year={2025},
      eprint={2506.15746},
      archivePrefix={arXiv},
      primaryClass={cs.NE},
      url={https://arxiv.org/abs/2506.15746}, 
}

@article{belkin2019reconciling,
  title={Reconciling modern machine-learning practice and the classical bias--variance trade-off},
  author={Belkin, Mikhail and Hsu, Daniel and Ma, Siyuan and Mandal, Soumik},
  journal={Proceedings of the National Academy of Sciences},
  volume={116},
  number={32},
  pages={15849--15854},
  year={2019},
  publisher={National Acad Sciences}
}

@article{kaplan2020scaling,
  title={Scaling laws for neural language models},
  author={Kaplan, Jared and McCandlish, Sam and Henighan, Tom and Brown, Tom B and Chess, Benjamin and Child, Rewon and Gray, Scott and Radford, Alec and Wu, Jeffrey and Amodei, Dario},
  journal={arXiv preprint arXiv:2001.08361},
  year={2020}
}

@article{hoffmann2022training,
  title={Training compute-optimal large language models},
  author={Hoffmann, Jordan and Borgeaud, Sebastian and Mensch, Arthur and Buchatskaya, Elena and Cai, Trevor and Rutherford, Eliza and Casas, Diego de Las and Hendricks, Lisa Anne and Welbl, Johannes and Clark, Aidan and others},
  journal={arXiv preprint arXiv:2203.15556},
  year={2022}
}

@inproceedings{brown2020language,
  title={Language models are few-shot learners},
  author={Brown, Tom and Mann, Benjamin and Ryder, Nick and Subbiah, Melanie and Kaplan, Jared D and Dhariwal, Prafulla and Neelakantan, Arvind and Shyam, Pranav and Sastry, Girish and Askell, Amanda and others},
  booktitle={Advances in Neural Information Processing Systems},
  volume={33},
  pages={1877--1901},
  year={2020}
}

@article{wang2020generalizing,
  title={Generalizing from a few examples: A survey on few-shot learning},
  author={Wang, Yaqing and Yao, Quanming and Kwok, James T and Ni, Lionel M},
  journal={ACM computing surveys (csur)},
  volume={53},
  number={3},
  pages={1--34},
  year={2020},
  publisher={ACM New York, NY, USA}
}

@inproceedings{carlini2021extracting,
  title={Extracting training data from large language models},
  author={Carlini, Nicholas and Tramer, Florian and Wallace, Eric and Jagielski, Matthew and Herbert-Voss, Ariel and Lee, Katherine and Roberts, Adam and Brown, Tom and Song, Dawn and Erlingsson, Ulfar and others},
  booktitle={30th USENIX Security Symposium (USENIX Security 21)},
  pages={2633--2650},
  year={2021}
}

@inproceedings{finn2017model,
  title={Model-agnostic meta-learning for fast adaptation of deep networks},
  author={Finn, Chelsea and Abbeel, Pieter and Levine, Sergey},
  booktitle={International conference on machine learning},
  pages={1126--1135},
  year={2017},
  organization={PMLR}
}

@inproceedings{snell2017prototypical,
  title={Prototypical networks for few-shot learning},
  author={Snell, Jake and Swersky, Kevin and Zemel, Richard},
  booktitle={Advances in neural information processing systems},
  volume={30},
  year={2017}
}

@article{turing1952chemical,
  title={The chemical basis of morphogenesis},
  author={Turing, Alan Mathison},
  journal={Philosophical Transactions of the Royal Society of London. Series B, Biological Sciences},
  volume={237},
  number={641},
  pages={37--72},
  year={1952},
  publisher={The Royal Society London}
}

@article{levin2012morphogenetic,
  title={Morphogenetic fields in embryogenesis, regeneration, and cancer: non-local control of complex patterning},
  author={Levin, Michael},
  journal={Biosystems},
  volume={109},
  number={3},
  pages={243--261},
  year={2012},
  publisher={Elsevier}
}

@article{nichele2017deep,
  title={Deep learning with cellular automaton-based reservoir computing},
  author={Nichele, Stefano and Molund, Andreas},
  journal={Complex Systems},
  volume={26},
  number={4},
  pages={319--340},
  year={2017},
  url={https://www.complex-systems.com/abstracts/v26_i04_a03/}
}

@article{mordvintsev2020growing,
  title={Growing neural cellular automata},
  author={Mordvintsev, Alexander and Randazzo, Ettore and Niklasson, Eyvind and Levin, Michael},
  journal={Distill},
  volume={5},
  number={2},
  pages={e23},
  year={2020}
}

@article{variengien2021towards,
  title={Towards robust and generalizable representations in neural cellular automata},
  author={Variengien, Alexandre and Najarro, Elias and Risi, Sebastian},
  journal={arXiv preprint arXiv:2111.00287},
  year={2021}
}

@article{gilpin2019cellular,
  title={Cellular automata as convolutional neural networks},
  author={Gilpin, William},
  journal={Physical Review E},
  volume={100},
  number={3},
  pages={032402},
  year={2019},
  publisher={APS}
}

@article{randazzo2020self,
  title={Self-classifying MNIST digits with neural cellular automata},
  author={Randazzo, Ettore and Mordvintsev, Alexander and Niklasson, Eyvind and Levin, Michael and Greydanus, Sam},
  journal={Distill},
  volume={5},
  number={8},
  pages={e27},
  year={2020}
}

@article{yosinski2014how,
  title={How transferable are features in deep neural networks?},
  author={Yosinski, Jason and Clune, Jeff and Bengio, Yoshua and Lipson, Hod},
  journal={NeurIPS},
  volume={27},
  year={2014},
  url={https://arxiv.org/abs/1411.1792}
}

@misc{zhuang2020comprehensivesurveytransferlearning,
      title={A Comprehensive Survey on Transfer Learning}, 
      author={Fuzhen Zhuang and Zhiyuan Qi and Keyu Duan and Dongbo Xi and Yongchun Zhu and Hengshu Zhu and Hui Xiong and Qing He},
      year={2020},
      eprint={1911.02685},
      archivePrefix={arXiv},
      primaryClass={cs.LG},
      url={https://arxiv.org/abs/1911.02685}, 
}

@inproceedings{nakkiran2020deep,
  title={Deep Double Descent: Where Bigger Models and More Data Hurt},
  author={Nakkiran, Preetum and Kaplun, Gal and Bansal, Yamini and Yang, Tristan and Barak, Boaz and Sutskever, Ilya},
  booktitle={International Conference on Learning Representations (ICLR)},
  year={2020},
  url={https://arxiv.org/abs/1912.02292}
}

@inproceedings{feldman2020does,
  title={Does learning require memorization? a short tale about a long tail},
  author={Feldman, Vitaly},
  booktitle={Proceedings of the 52nd Annual ACM SIGACT Symposium on Theory of Computing (STOC)},
  pages={954--959},
  year={2020},
  url={https://arxiv.org/abs/1906.05271}
}

@inproceedings{sudhakaran2021growing,
  title={Growing 3D Artefacts and Functional Machines with Neural Cellular Automata},
  author={Sudhakaran, Shyam and Grbic, Djordje and Li, Siyan and Katona, Adam and Najarro, Elias and Glanois, Claire and Risi, Sebastian},
  booktitle={Proceedings of the 2021 Conference on Artificial Life (ALIFE 2021)},
  year={2021},
  url={https://arxiv.org/abs/2103.08737}
}

@inproceedings{earle2022illuminating,
  title={Illuminating diverse neural cellular automata for level generation},
  author={Earle, Sam and Snider, Justin and Fontaine, Matthew C and Nikolaidis, Stefanos and Togelius, Julian},
  booktitle={Proceedings of the Genetic and Evolutionary Computation Conference (GECCO)},
  pages={68--76},
  year={2022},
  url={https://arxiv.org/abs/2109.05489}
}

@misc{palm2022variational,
      title={Variational Neural Cellular Automata}, 
      author={Rasmus Berg Palm and Miguel González-Duque and Shyam Sudhakaran and Sebastian Risi},
      year={2022},
      eprint={2201.12360},
      archivePrefix={arXiv},
      primaryClass={cs.NE},
      url={https://arxiv.org/abs/2201.12360}, 
}

@article{niklasson2021textures,
  title={Self-Organising Textures},
  author={Niklasson, Eyvind and Mordvintsev, Alexander and Randazzo, Ettore and Levin, Michael},
  journal={Distill},
  volume={6},
  number={2},
  pages={e00027-003},
  year={2021},
  url={https://distill.pub/2021/selforg-textures/}
}

@inproceedings{walker2022physical,
  title={Physical Neural Cellular Automata for 2D Shape Classification},
  author={Walker, Kathryn and Palm, Rasmus Berg and Garcia, Rodrigo Moreno and St{\o}y, Kasper and Risi, Sebastian},
  booktitle={Proceedings of the Artificial Life Conference (ALIFE)},
  year={2022},
  url={https://arxiv.org/abs/2203.07548}
}

@inproceedings{Cisneros_2020, series={ALIFE 2020},
   title={Visualizing computation in large-scale cellular automata},
   url={http://dx.doi.org/10.1162/isal_a_00277},
   DOI={10.1162/isal_a_00277},
   booktitle={The 2020 Conference on Artificial Life},
   publisher={MIT Press},
   author={Cisneros, Hugo and Sivic, Josef and Mikolov, Tomas},
   year={2020},
   pages={239–247},
   collection={ALIFE 2020} }
\end{document}